%% file: template.tex
\documentclass[10pt,twocolumn,letterpaper]{article}
\usepackage{subcaption}
\usepackage{gensymb}
\usepackage{iccv}
\usepackage{times}
\usepackage{epsfig}
\usepackage{graphicx}
\usepackage{amsmath}
\usepackage{amssymb}
\usepackage{authblk}

\newcommand{\para}[1]{\noindent{\bf #1}}

\newcommand{\compactsection}[1]{\vspace{-2mm} \section{#1} \vspace{-1mm}}
\newcommand{\compactsubsection}[1]{\vspace{-0mm} \subsection{#1} \vspace{-0mm}}
\usepackage{textcomp}

\usepackage[pagebackref=true,breaklinks=true,letterpaper=true,colorlinks,bookmarks=false]{hyperref}

\iccvfinalcopy 

\def\iccvPaperID{2042} 

\ificcvfinal\fi

\begin{document}

\title{Learning Dense Facial Correspondences in Unconstrained Images}

\author[1,3]{Ronald Yu
\thanks{ ronaldyu@usc.edu}}
\author[1,3]{Shunsuke Saito
\thanks{shunsuke.saito16@gmail.com}}
\author[2]{Haoxiang Li
\thanks{haoxli@adobe.com}}
\author[2]{Duygu Ceylan
\thanks{ceylan@adobe.com}}
\author[1,3,4]{Hao Li\thanks{hao@hao-li.com}}
\affil[1]{University of Southern California}\affil[2]{Adobe Research}\affil[3]{Pinscreen}\affil[4]{USC Institute for Creative Technologies}

\maketitle

\begin{abstract}
\vspace{-3mm}
We present a minimalistic but effective neural network that computes dense facial correspondences in highly unconstrained RGB images. Our network learns a per-pixel flow and a matchability mask between 2D input photographs of a person and the projection of a textured 3D face model. To train such a network, we generate a massive dataset of synthetic faces with dense labels using renderings of a morphable face model with variations in pose, expressions, lighting, and occlusions. We found that a training refinement using real photographs is required to drastically improve the ability to handle real images. When combined with a facial detection and 3D face fitting step, we show that our approach outperforms the state-of-the-art face alignment methods in terms of accuracy and speed. By directly estimating dense correspondences, we do not rely on the full visibility of sparse facial landmarks and are not limited to the model space of regression-based approaches. We also assess our method on video frames and demonstrate successful per-frame processing under extreme pose variations, occlusions, and lighting conditions. Compared to existing 3D facial tracking techniques, our fitting does not rely on previous frames or frontal facial initialization and is robust to imperfect face detections.
\vspace{-3mm}
\end{abstract}
\input{intro}
\input{related}
\input{algorithm}

\input{experiments}

\input{conclusion}
{\small
\bibliographystyle{ieee}
\bibliography{egbib}
}

\end{document}

%% file: intro.tex
\compactsection{Introduction}

By introducing 3D facial alignment techniques that can process images in the wild, it is possible to improve the performance of facial recognition methods~\cite{Blanz:2003:FRB,PaysanKARV09,ChuRC14,HuYCDCKR16}; compelling 3D face models of a person can be generated for gaming and virtual reality applications~\cite{Blanz:1999:MMS,Ichim:2015:DAC,li2015facial,cao2016real,SaitoWHNL16,olszewski2016high}; and an accurate tracking model can be initialized for real-time facial performance capture and animation~\cite{Saragih:2011:DMF,cao2014displaced,thies2016face,saito2016realtime}. Most of the techniques rely on a robust detection of sparse facial landmarks (eyes, nose, lips, etc.) and tend to perform best when most features are visible, front-facing, and free from occlusions or challenging lighting conditions. In many applications, such as video facial analytics or driver attention monitoring, these assumptions do not hold since the subjects are recorded in a fully unconstrained environment. 

With the recent advancement of deep learning techniques, highly robust regression methods have emerged that can successfully fit a 3D face model for extremely difficult cases, such as side views of a face or occlusions by hair. 
State-of-the-art methods are based on regression~\cite{Zhu_2016_CVPR,jourabloo2016large} and directly regress the shape parameters of a 3D morphable model (3DMM)~\cite{Blanz:1999:MMS} and expression coefficients~\cite{cao2014facewarehouse} from an image using cascaded network structures. While achieving impressive accuracies on several challenging benchmark datasets, they still tend to perform poorly in extreme real-case scenarios, as demonstrated in this paper. Not only is such approach limited to variations defined by the face model space, but its performance relies on a perfectly tight facial bounding box detection. Despite significant progress, even the cutting edge face detectors~\cite{HuR16} cannot guarantee a clean face localization and cropping for extreme images. 

Instead of a regression method, we propose an alternative deep learning approach that estimates dense pixel-wise correspondences between the input image and a 3DMM model along with a \emph{matchability mask}, which defines which pixels belong to the face and have valid correspondences. We perform dense correspondence estimation by predicting a per-pixel 2D flow vector between the input image and a synthetic rendering of a 3DMM. Once the correspondences are established, we fit a 3DMM to the input using available correspondences. Compared to sparse landmark detection techniques, dense correspondences provide more robust constraints, since any part of the face can be used for matching, and our predicted matchability mask helps to distinguish non-visible parts of the face. Furthermore, our 2D flow computation is less sensitive to clean bounding box estimations during an initial face detection as oppose to existing regression approaches.

Inspired by the recent work of~\cite{Zhou_2016_CVPR}, we train a simple encoder-decoder network using synthetically generated 3DMMs with variations in pose, shape, appearance, and lighting, and simulate occlusions using random box renderings. Since every face of the 3DMM have consistent mesh topologies, the dense labels are automatically present. We further refine the training using real photographs with corresponding face models obtained from the regression technique of~\cite{Zhu_2016_CVPR}, and predict the flow between the input image and a statistical mean face. We show that by combining our dense correspondence computation with a subsequent face fitting step, we can perform comparably with the current state-of-the-art face alignment techniques on difficult images in terms of accuracy and are significantly faster on public datasets. Furthermore, we demonstrate highly effective pose estimations and 3D face fittings on extremely challenging images and videos. 

%% file: related.tex
\compactsection{Related Work}

\paragraph{2D Face Alignment.} Facial alignment for images and videos has attracted a lot of attention from the research community due to its wide range of applications. 2D facial alignment approaches aim to localize a set of fiducial points in the face. Classical approaches include the Active Appearance Models (AAM)~\cite{Cootes98activeappearance,matthews2004active,saragih2007nonlinear,tzimiropoulos2013optimization} and Constrained Local Models (CLM)~\cite{Cristinacce:2008:AFL,Saragih:2011:DMF,asthana2013robust}. Another common approach is to learn regression functions that map hand-crafted image features to 2D landmark positions directly~\cite{valstar2010facial,xiong2013supervised,cao2014face,asthana2014incremental,ren2014face,kazemi2014one,lee2015face,Xiong2015,zhu2015face}. With the recent success of deep learning methods, several approaches have replaced the use of hand-crafted features with a convolutional neural network~\cite{sun2013deep,zhou2013extensive,zhang2014facial,peng2016recurrent}. While such purely 2D methods have shown impressive results especially for frontal and non-occluded faces, modeling of occlusions has been mostly avoided. More recent approaches~\cite{burgos2013robust, jia2014structured, ghiasi2014occlusion, yang2015robust, saito2016realtime} have attacked this problem by introducing occlusion variation in the training data.  Handling large pose variations under difficult illumination conditions, however, still remains challenging for 2D methods. To handle large pose variations, several approaches have proposed to use multiple shape models for different views~\cite{cootes2002view, zhu2012face, yu2013pose}. However, due to the requirement to test all these possible views, such methods are computationally very expensive.

\vspace {-4mm}
\paragraph{3D Face Alignment.} In the context of 3D facial alignment, earlier works have focused on optimization based methods that minimize the difference between the input image and the model appearance~\cite{Blanz:1999:MMS,romdhani2005estimating}. As in the case of 2D facial alignment, an alternative approach is to regress the parameters of a 3D face model based on image features around landmark points~\cite{cao2014displaced,jeni2015dense,Jourabloo2015}. More recent methods have focused on performing this regression with neural networks, specifically with cascaded~\cite{Zhu_2016_CVPR,jourabloo2016large,liu2016joint} or very deep network structures~\cite{tran2016regressing,laine2016facial}. While such methods achieve impressive results on challenging datasets, the main drawback is being limited to the shape space represented by the utilized 3D morphable model. We present an alternative approach of predicting dense correspondences between the input image and the face model, which can potentially be propagated to any 3D morphable model with little effort. We provide extensive comparisons between our alternative approach and the recent 3D facial alignment methods and show that our method outperforms both in terms of accuracy and speed (see Section~\ref{sec:exp}). 

\vspace{-5mm}
\paragraph{Dense Correspondence Estimation.} The problem of dense correspondence estimation has been mainly investigated in the form of \emph{optical flow estimation} for tracking purposes~\cite{horn1981determining,black1996robust, baker2011database, brox2011large}. To compute dense correspondences between different scenes and different instances of an object category, energy minimization approaches that match hand-crafted features with additional smoothness priors have been proposed~\cite{Barnes:2010,liu2011sift,kim2013deformable,bristow2015dense}. Such approaches have been further improved by either jointly solving for co-segmentation~\cite{taniai2016joint} or analyzing collections of images~\cite{zhou2015flowweb}. Last but not least, recent methods have explored the power of neural networks for predicting dense optical flow~\cite{weinzaepfel2013deepflow, dosovitskiy2015flownet, ilg2016flownet} and correspondences~\cite{Zhou_2016_CVPR}. These methods have shown impressive results that motivate us to explore the power of predicting dense flow to tackle the problem of 3D facial alignment. \cite{guler2016densereg} has also explored using dense correspondences for the purpose of 3D facial landmark alignment. We show that dense correspondences provide robust constraints for 3D face fitting under large pose and illumination conditions since they enable any part of the face to be utilized for matching.

\vspace{-0.5mm}

%% file: algorithm.tex



\vspace{-1mm}
\compactsection{Proposed Method}
\vspace{-1mm}
\begin{figure*}[!h]
\centering
\includegraphics[width=\linewidth]{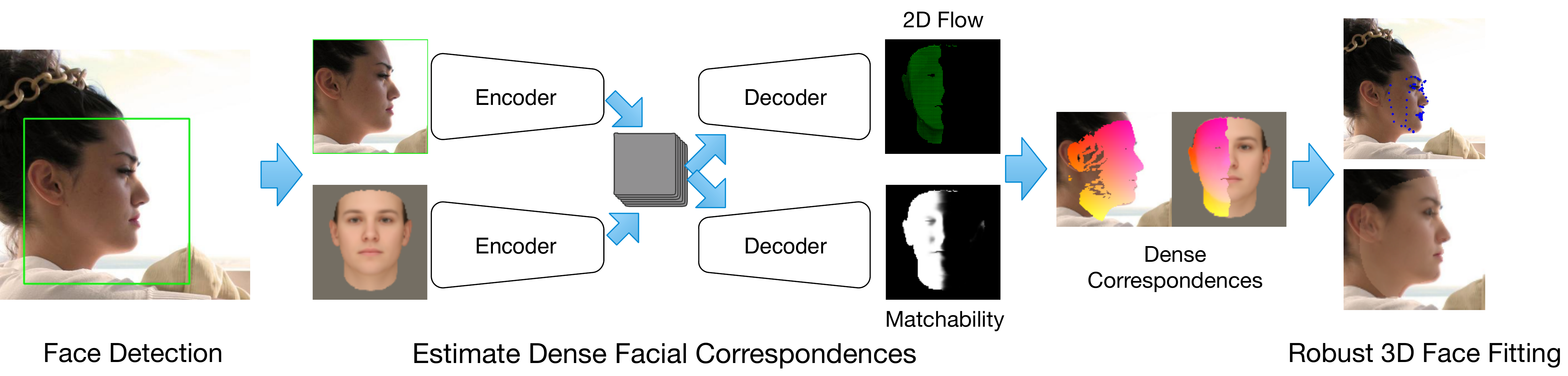}
\caption{Given an input image with a detected face, we propose an encoder-decoder architecture that predicts Dense Facial Correspondences between the input image and a 3D morphable face model. These correspondences are estimated as 2D flow between the input image and a synthetic rendering of a frontal, mean face. We also predict a matchability mask which indicates which correspondences are valid or not. Using such correspondences, we can perform 3D face alignment even in very challenging cases of large pose and illumination variation.}
\label{fig:workflow-testing}
\vspace{-5mm}
\end{figure*}
\vspace{0.5mm}
\compactsubsection{Overview}
\vspace{-1mm}
Our approach takes a source image and outputs per-pixel correspondences between the source image and a 3D morphable model (3DMM). Since correspondences are well-defined only on regions of the face that are visible in the source image, we also output a \emph{matchability mask} that predicts the probability of each correspondence being valid or not. We perform dense correspondence estimation by predicting a per-pixel 2D flow between the source image and a synthetic rendering of the 3DMM depicting a frontal mean face. Both 2D flow and the matchability are predicted by a convolutional neural network (Section~\ref{sec:pred}). Valid 2D correspondences are easily translated to 2D-3D correspondences since each pixel in the synthetic rendering is directly associated with the 3DMM. Such 2D-3D correspondences are then used to guide the alignment of the 3DMM to the source image (Section~\ref{face3D}).
 This pipeline is illustrated in Figure~\ref{fig:workflow-testing}. We next discuss each stage in more detail.

%
%

\compactsubsection{3D Morphable Model}
We use the 3D morphable model (3DMM) proposed by Blanz and Vetter~\cite{Blanz:1999:MMS}. Each 3D face, $S$, is represented as:
\begin{equation}
S = \bar{S} + A_{id}\alpha_{id} + A_{exp}\alpha_{exp},
\end{equation}
where $\bar{S}$ is the mean 3D face, $A_{id}$ and $A_{exp}$ are the basis for the identity and the expression respectively. $\alpha_{id}$ and $\alpha_{exp}$ denote the parameters for the identity and the expression basis. Moreover, in order to project a 3D face $S$ to a 2D image we use perspective projection:
\begin{equation}
S_{2D} = \Pi_f  (\mathbf{R} (\bar{S} + A_{id}\alpha_{id} + A_{exp}\alpha_exp) + \mathbf{t}),
\label{eq:proj}
\end{equation}
where $\Pi_f$ is the projection operator that depends on the focal length, $f$, and the principal point defined to be the center of the image. $\mathbf{R}$ and $\mathbf{t}$ denote the rotation and the translation components of the pose. Thus, aligning the 3DMM with an image is equivalent to finding the set of parameters $(f, \mathbf{R}, \mathbf{t}, \alpha_{id}, \alpha_{exp})$ that minimizes an alignment error as described in Section~\ref{face3D}.

\begin{figure}[t]
\begin{center}
   \includegraphics[width=0.24\linewidth]{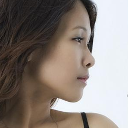}
   \includegraphics[width=0.24\linewidth]{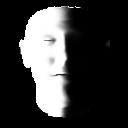}
   \includegraphics[width=0.24\linewidth]{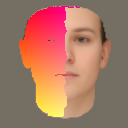}
   \includegraphics[width=0.24\linewidth]{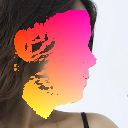}
   \includegraphics[width=0.24\linewidth]{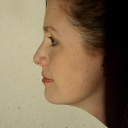}
   \includegraphics[width=0.24\linewidth]{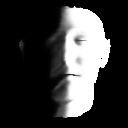}
   \includegraphics[width=0.24\linewidth]{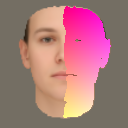}
   \includegraphics[width=0.24\linewidth]{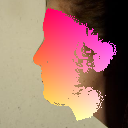}
   \includegraphics[width=0.24\linewidth]{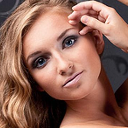}
   \includegraphics[width=0.24\linewidth]{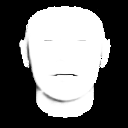}
   \includegraphics[width=0.24\linewidth]{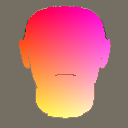}
   \includegraphics[width=0.24\linewidth]{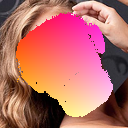}
\end{center}
   \caption{Sample visualizations of dense correspondences between our input and our template image for two inputs. The first image is the input. The second image is the matchability map that indicates which pixels on the template are matchable. Note that we do not explicitly segment out external occlusions and occluded areas are also considered matchable as long as they reside in the face shape. The two color maps on the right show the dense correspondence between the two images. The holes found at the edges of the dense correspondences in the third column indicate that the pixel cannot be matched and does not have a flow to our frontal facing template.} \label{dense}
\label{fig:architecture}
\end{figure}

\compactsubsection{Dense Correspondence Prediction}
\label{sec:pred}
Given a source image, $I_s$, and a rendering of the 3DMM showing a frontal, mean 3D face, which we call the target image $I_t$, we propose a network architecture to predict a per-pixel 2D flow from $I_s$ to $I_t$ along with a matchability mask. For each pixel location $\mathbf{p}_s=(x,y)$ in $I_s$, the 2D flow $\mathbf{F}_{s,t}(x,y) = ({\Delta}x, {\Delta}y)$ maps $\mathbf{p}_s$ to the location $\mathbf{q}_t=(x+{\Delta}x, y+{\Delta}y)$ in $I_t$, predicting that $\mathbf{p}_s$ and $\mathbf{q}_t$ are semantic correspondences. Since the 2D flow is well defined only for parts of the face visible in $I_s$ and $I_t$, we also predict a matchability score $m_{s,t}(\mathbf{p}_s) \in [0,1]$, where $m_{s,t}(\mathbf{p}_s) = 1$ if $p_s$ has a valid correspondence in $I_t$. We note that we first perform face detection in the source image and predict the flow on the cropped image based on the detection result.
A visualization of our  network output can be seen in Figure ~\ref{dense}

\begin{figure*}[h]
\begin{center}
\includegraphics[width=\textwidth]{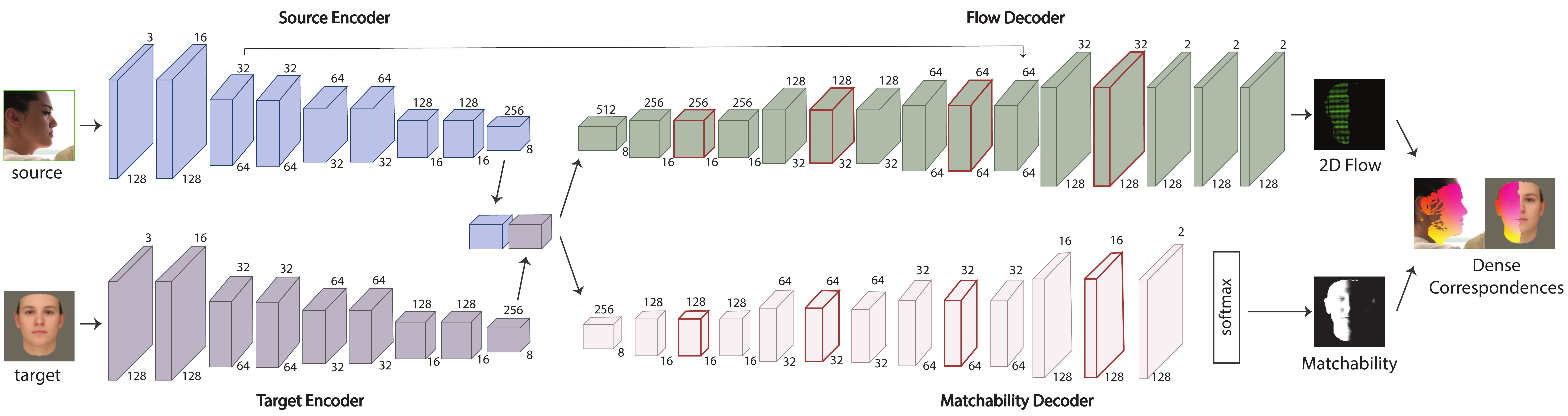}
\caption{Our network architecture.}
\label{fig:arch}
\end{center}
\vspace{-8mm}
\end{figure*}

Our network architecture for predicting the 2D flow and the matchability generally follows the architecture recently proposed by Zhou et al.~\cite{Zhou_2016_CVPR}. Specifically, we use two encoder branches that take the source and target images as input respectively. The output of these encoders are concatenated and provided as input to two decoder branches. The \emph{flow decoder} outputs a 2-channel feature map with the same size as the source image, where the channels specify the 2D flow $({\Delta}x,{\Delta}y)$ for each pixel. The \emph{matchability decoder}, on the other hand, outputs a single channel feature map with the same size as the source image representing the probability of each pixel being matchable. 

The encoders consist of 8 convolutional layers, each followed by a ReLU layer. Every second convolution layer has a stride of 2 in order to decrease the spatial dimension by half. Each decoder begins with four triplets of convolution layers. In each triplet, the third layer is a deconvolutional layer with a stride of 2. The first and third layer of each triplet is followed by an ReLU layer. For the flow decoder, the triplets are followed by three additional convolutional layers with an ReLU in between. For the matchability decoder, the triplets are followed by a single convolutional layer and a softmax function that classifies each pixel as matchable or not matchable. Details of the network architecture can be seen in Figure ~\ref{fig:arch}.
\\

\para{Loss function}. Given a source and a target image $I_s$ and $I_t$, for each pixel $(x,y)$ in $I_s$ we denote by $\mathbf{F}(x,y)$ and $m(x,y)$ the 2D flow and matchability predictions and their ground truths by $\widetilde{\mathbf{F}}(x,y)$ and $\widetilde{m}(x,y)$ respectively. We train the network to minimize the following loss $\mathbf{L}(I_s, I_t)$:
\begin{eqnarray*}
    \mathbf{L}(I_s, I_t) &=& \sum_{x,y} \widetilde{m}(x,y) ||\mathbf{F}(x,y) - \widetilde{\mathbf{F}}(x,y)||^2 \\
    &+& \lambda \sum_{x,y} \mathbf{L}_C (m(x,y), \widetilde{m}(x,y)),
\end{eqnarray*}
where $\mathbf{L}_C$ denotes the cross-entropy loss and $\lambda$ is a hyper-parameter.
\\

\para{Training procedure} In order to train the proposed network architecture, we need access to images where ground truth correspondences with a 3DMM are available. We use the recently released large-pose 300W (300W-LP) dataset~\cite{Zhu_2016_CVPR} which provides the parameters of a 3DMM fitting each image in the dataset. Given the pose and the 3DMM parameters, we project the 3D face to the input image using Equation~\ref{eq:proj}. For each pixel $p_s$ in the input image, we identify the $uv-$coordinate of the 3DMM surface point that projects to it. Then, we find the pixel $q_t$ in the rendering of the frontal, mean face that has the most similar $uv-$ coordinate. If the distance between the $uv-$coordinates is less than a threshold ($0.015$ in our experiments), we define a ground truth correspondence between the pixel $p_s$ in the input image and $q_t$ in the rendering of the frontal, mean template.

We observe that training the network from scratch directly by feeding ``in-the-wild'' real face images does not converge. We assume this is due to the large appearance variations as well as the noisy ground-truth annotations. To overcome this challenge, we propose to use a large scale synthetic data in a pre-training process where we learn dense correspondences between random pairs of synthetic faces. Specifically, using the 3DMM we generate random pairs of synthetic renderings showing faces with varying identity, expression, pose, lighting, and occlusion (see Figure~\ref{fig:syntheticfaces}). Since both images in the pair are generated from the 3DMM, we have direct access to perfect ground truth correspondences and matchability masks. Note that although our framework is robust to external occlusions and  able to detect self-occlusions in our matchability mask, we do not explicitly
segment out external occlusions such as in \cite{saito2016realtime} due to limitations in our training data. During this pre-training stage, since we provide both of the encoders with synthetic renderings we share their weights. After convergence, the network accurately estimates dense correspondences between two synthetic faces, even with extreme pose and lighting.

We next fine-tune our network on the 300W-LP dataset~\footnote{300W-LP also contains synthetic faces, but with more realistic texture and background compared with our synthetic faces.}. We fix the input to the target encoder branch to be the frontal, mean face rendering while the input to the source branch are the real face images. Thus, in this stage the two encoder branches no longer share weights. 

Although the input to one of the encoder branches is fixed, we note that this input branch is essential to our framework.
If we only use a one-encoder-branch network architecture during the pre-training stage, 
training with synthetic data may adversely lead to overfitting to the appearance of our synthetic images. To address this, we have two input branches take an image pair to predict the flow to guide the network to learn the correspondences instead of memorizing the appearance. We then fine-tune the network with real data while keeping one input fixed. In our experiments, we observed that the training does not converge when discarding the constant branch since the task would not align well to the pre-training setting. For this reason, we keep a constant encoder branch in our final model.

\compactsubsection{3DMM Alignment}
\label{face3D}
Once our network is trained, during test time, given a single input image we first predict dense correspondences and the matchability mask with respect to the rendering of the frontal, mean face template. We filter our correspondences with low matchability scores and translate the remaining correspondences to 2D-3D correspondences between the input image and the 3DMM. We use such 2D-3D correspondences to fit the 3DMM to the input image to further refine our results. 

Given a set of $(p^i, q^i)$ correspondences where $p^i$ denotes a pixel in the input image and $q^i$ denotes its corresponding vertex in the 3DMM, we minimize the following energy function defined over the parameters $\mathcal{X} = (f, \mathbf{R}, \mathbf{t}, \alpha_{id}, \alpha_{exp})$:

\begin{eqnarray}
E(\mathcal{X}) &=& E_{data}(\mathcal{X}) + E_{reg}(\mathcal{X}),  \\
E_{data}(\mathcal{X}) &=& \sum_i w_i \|p^i - \Pi_f(\mathbf{R}S_{q^i}+ \mathbf{t})\|^2, \nonumber \\
E_{reg}(\mathcal{X}) &=& w_{id}\sum_{id,i} (\frac{\alpha_{id,i}}{\sigma_{id, i}})^2 + w_{exp}\sum_{exp,i} (\frac{\alpha_{exp,i}}{\sigma_{exp, i}})^2 \nonumber,
\end{eqnarray}

where $E_{data}$ measures the error between the projected 3D face points and their corresponding 2D points and $E_{reg}$ is a statistical prior over the identity and the expression blendshapes~\cite{SaitoWHNL16} and we set $w_{id}=2.5\times 10^{-5},w_{exp}=1000$. $S_{q^i} = (\bar{S} + A_{id}\alpha_{id} + A_{exp}\alpha_{exp})_{q^i}$ denotes the position of the vertex $q_i$ in the 3D face defined by the parameters $(\alpha_{id}, \alpha_{exp})$. $w_i$ defines the weight of each correspondence and all weights are initialized to be equal. 
We iteratively solve for the parameters listed in Eq. 3 to minimize the L2 distance between the mesh\textquotesingle s projected vertices and their estimated pixel location based on the dense correspondence at each iteration.
This standard formulation is also used in many previous papers such as \cite{thies2016face}, \cite{saito2016realtime}, and \cite{cao2014displaced}.
Once the parameters of the 3DMM that aligns best with the input image are computed, we recover any missing correspondence and refine our predictions.

%% file: experiments.tex
\begin{figure}[h]
\begin{center}
\includegraphics[width=0.72\columnwidth]{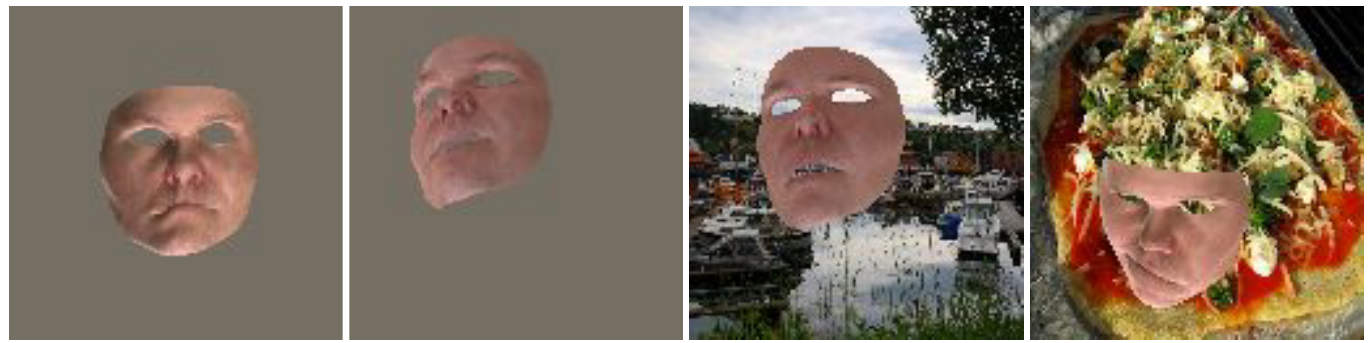}
\includegraphics[width=0.18\columnwidth]{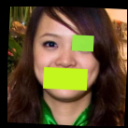}
\caption{We train our network first on a large set of synthetic data with variations in shape, expression, facial texture, and illumination. We further composite some of these renderings with real background images. We later synthesize occlusions onto real background images as seen in the image on the far right.}
\label{fig:syntheticfaces}
\end{center}
\vspace{-5mm}
\end{figure}

%
\compactsection{Experiments}
\label{sec:exp}
\vspace{-1mm}
\compactsubsection{Training Data}
\vspace{-1mm}
Our network operates on images of size $128 \times 128$. We train our network first on a large set of synthetic renderings of a 3D morphable model. To generate these renderings, we randomly sample different facial textures from the Chicago face database~\cite{Ma2015}, we sample the identity and expression parameters as well as the rotation and translation from a Gaussian distribution. We also randomly sample spherical harmonics values from a database of lighting environments and apply it to the face to generate a total of 200k renderings on gray background. We also composite an additional 200k renderings with random background images downloaded from the COCO dataset~\cite{Lin2014}(see Figure ~\ref{fig:syntheticfaces}). We use a total of 100k random pairs of source and target images with gray background and 100k random pairs of source and target images with real background for training. We use a batch size of 12 and learning rate of $1\mathrm{e}{-4}$ for roughly 2 epochs. One epoch takes roughly 6 hours.

\begin{figure*}[h]
\begin{center}
\includegraphics[width=\textwidth]{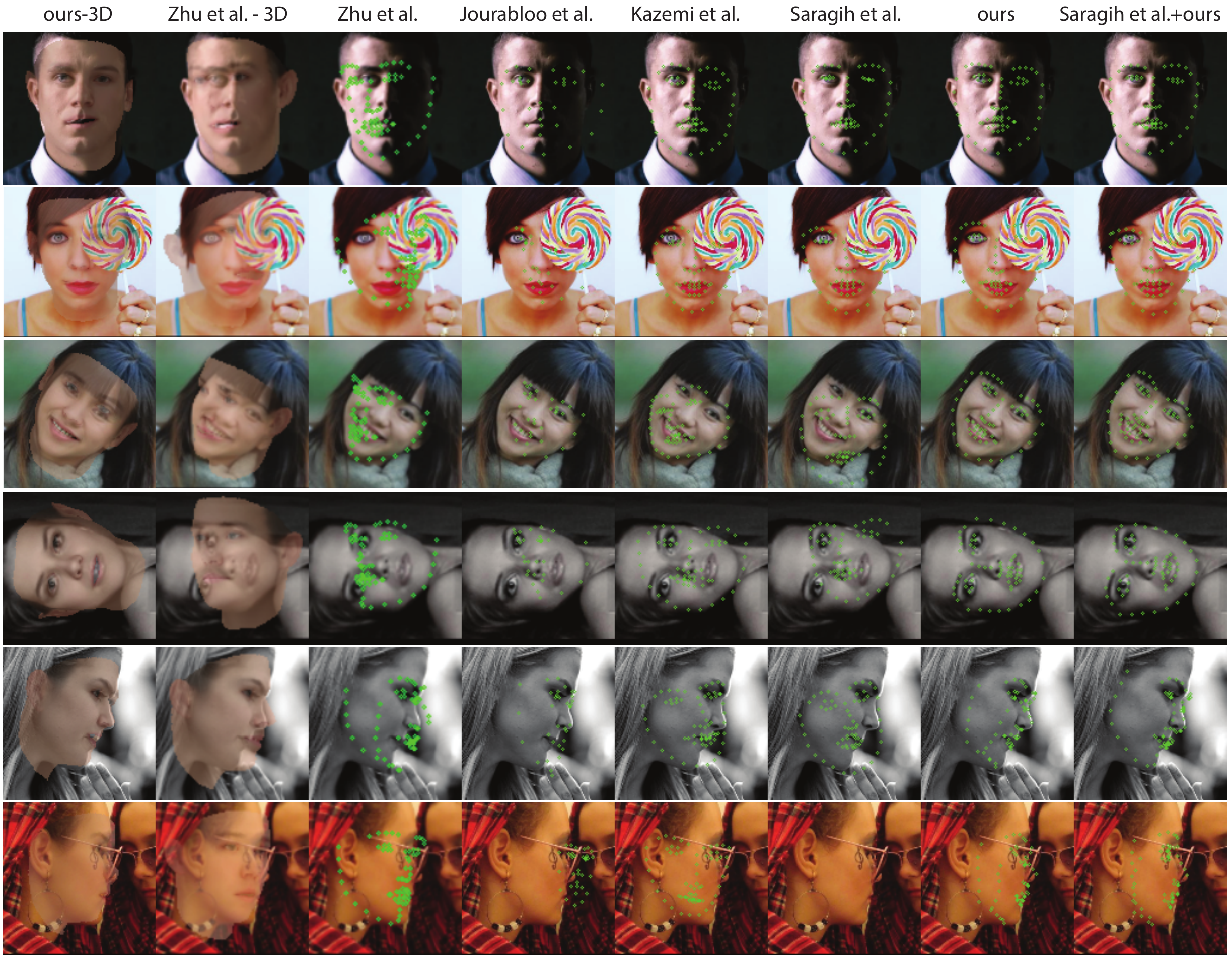}
\caption{We provide visual 2D and 3D facial alignment results on the AFLW2000 dataset~\cite{Zhu_2016_CVPR} using our method, the method of Zhu et al.~\cite{Zhu_2016_CVPR}, Jourabloo et al.~\cite{jourabloo2016large}, Kazemi et al.~\cite{kazemi2014one}, and Saragih et al.~\cite{saragih2007nonlinear}. We also show the results obtained by~\cite{saragih2007nonlinear} when initialized with our predictions.}
\label{fig:aflw}
\end{center}
\vspace{-8mm}
\end{figure*}

Once the network converges, we fine-tune it with the images from the 300W-LP dataset~\cite{Zhu_2016_CVPR}. We also perturb images from the 300W-LP dataset with 2D image-plane scale, translation, and rotation, and we also synthesize occlusions by drawing rectangles similar to the method of \cite{saito2016realtime}. An example of the rectangles we synthesize to simulate occlusion is seen in Figure  ~\ref{fig:syntheticfaces} . We first train with a learning rate of $1\mathrm{e}{-4}$ for about two epochs and then drop the learning rate by a factor of 10 and train for another epoch.
\compactsubsection{Qualitative Evaluations}
We evaluate the performance of our method both for 2D and 3D facial alignment on the recently released AFLW2000~\cite{Zhu_2016_CVPR} dataset of challenging and large pose images and show qualitative results in Figure~\ref{fig:aflw}. We provide comparisons with the recent methods that tackle the problem of face alignment under large pose variations~\cite{Zhu_2016_CVPR,jourabloo2016large} as well as state-of-the-art face trackers including Kazemi et al.~\cite{kazemi2014one} and the TVS implementation of Saragih et al.~\cite{saragih2007nonlinear}. Both of them have been widely deployed in the industry. 
Furthermore, we demonstrate the performance of combining our method with 2D face alignment method by initializing the face tracker with our predictions of 2D facial landmarks.
We observe that our method is more robust to heavy occlusions, large variations in illumination, translation, and image-axis rotation. Our method can also serve as a better starting point for 2D face alignment method such as 
Saragih et al.~\cite{saragih2007nonlinear} to significantly improve its performance.
%
%

We also report evaluations on extremely challenging images and video sequences captured in the wild
in the supplemental materials. 



\compactsubsection{Quantitative Evaluation}
In addition to qualitative results, we perform quantitative evaluations by measuring the accuracy of the 2D facial landmarks. 
Once a 3D face model is aligned to an input image using the estimated dense correspondences, we can identify the 2D facial landmarks from the annotated vertices on the 3D model. We then measure the normalized mean error (NMS)~\cite{Zhu_2016_CVPR} between the ground truth and predicted facial landmarks.

We evaluate our method on several challenging datasets such as the 68 landmarks on AFLW2000~\cite{Zhu_2016_CVPR}, 21 landmarks on AFLW~\cite{tugraz:icg:lrs:koestinger11b}, and 21 landmarks on AFLW-PIFA~\cite{Jourabloo2015}~\footnote{On AFLW-PIFA, the ground-truth annotations have 34 landmarks. But we are only clear about their definitions on a 3D face for 21 out of them.}. Since we did not have much training data with real images, we included images from the 300W Challenge dataset \cite{sagonas2013300}  and their synthesized side views in our training set, so we did not evaluate our method on this dataset. We compare our performance to state-of-the-art face alignment methods including Zhu et al. \cite{Zhu_2016_CVPR} and Jourabloo et al.~\cite{jourabloo2016large}. 
\begin{table}[htp]
\begin{center} 
\begin{tabular}{ c|c|c|c } 
Method&0 to 30& 30 to 60 & 60 to 90\\
\hline
RCPR~\cite{burgos2013robust} & 4.26 & 5.96 & 13.18\\
\hline
ESR~\cite{cao2014face} & 5.60 & 6.70 &12.67\\
\hline
SDM~\cite{Xiong2015}& \textbf{3.67} & \textbf{4.94} & 9.76\\
\hline
Zhu et al. \cite{Zhu_2016_CVPR} & 3.78 & \textbf{4.54} & \textbf{7.93}\\
\hline
Our Method & \textbf{3.62} & 6.06 & \textbf{9.56}  \\
\hline
\end{tabular}
\vspace{0.5em}
\caption{Performance evaluation on AFLW2000 (68 landmarks): we report the NMS for faces in small ($[0,30]$), medium($[30,60]$), and large($[60,90]$) pose with respect to the yaw angles. 
The top two results in each category are highlighted in bold.} \label{aflw2000_1}
\end{center}
\vspace{-2mm}
\end{table}

In Table~\ref{tab:aflw}, we report our results on the visible landmarks on the complete AFLW~\cite{tugraz:icg:lrs:koestinger11b} along with the accuracy achieved by Zhu et al.~\cite{Zhu_2016_CVPR}, which has been shown to outperform previous existing methods on this dataset. 

\begin{figure}[h]
\begin{center}
\includegraphics[width=0.3\linewidth]{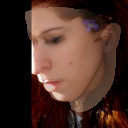}
\includegraphics[width=0.3\linewidth]{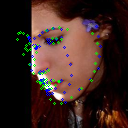}
\caption{The problem of evaluation against invisible and contour landmarks: we show a typical large-pose face image in AFLW2000 and its landmarks. 
On the left we show our 3D face fitting result. 
On the right we show the projected landmarks of our fitting are in green and ground truth landmarks are in blue. 
While the fitting is decent, the NMS over 68 landmarks is as high as 9.53.}
\label{ortho}
\end{center}
\vspace{-3mm}
\end{figure}

In Table ~\ref{aflw2000_1} we show our performance with NMS overall of the ground truth 68 landmarks in AFLW2000.
One issue of this setting is that the exact locations of invisible and contour landmarks is unclear~\cite{zhu2015high} and subjective. Moreover, their 2D projections may vary with different fitting and projection methods. 
Zhu et al.~\cite{zhu2015high} propose ``landmark marching'' to address this problem. But still the definitions of contour landmarks are arguable.
In Figure~\ref{ortho}, we observe that a decent-quality fitting can have a high NMS due to the subjective nature of the contour and invisible landmarks.
A strong motivation for detecting the contour and invisible landmarks is to help 3D face fitting. However, in our method, we are able to get accurate 3D face fitting from dense correspondences without relying 
on invisible landmarks and the need of defining contour landmarks.

To better understand the performance of our method, we exclude the contour and invisible landmarks and evaluate the NMS over only the inner and visible landmarks.
For each face image, the ground-truth visibility of a landmark can be obtained from its ground-truth 3D face provided in the AFLW2000 dataset.
We report our results in Table~\ref{aflw2000_2}. We observe that our method consistently achieves comparable performance with the state-of-the-art for faces in small, medium, and large pose.



%
\begin{table}[h!]
\begin{center} 
\begin{tabular}{ c|c|c } 
Pose (Yaw Angle) & Zhu et al \cite{Zhu_2016_CVPR}  &Our Method\\ 
	\hline
Small [0\degree-30\degree] & 4.30   & 3.14 \\ 
	\hline
Medium [30\degree-60\degree] & 4.41   & 3.84 \\ 
	\hline
Large [$>$60\degree] & 6.68  & 5.53 \\ 
	\hline
All Images & 4.60  & 3.58 \\ 
\hline
\end{tabular}
\vspace{0.5em}
\caption{Performance evaluation on AFLW2000 for visible inner landmarks: we report the NMS for faces in small ($[0,30]$), medium($[30,60]$), large($[60,90]$) pose with respect to the yaw angles, and across all the images.} \label{aflw2000_2}
\end{center}
\vspace{-2mm}
\end{table}
%
%
%

%
\begin{table}
\begin{center} 
\begin{tabular}{ c|c|c } 
Pose (Yaw Angle) & Zhu et al \cite{Zhu_2016_CVPR}  &Our Method\\ 
	\hline
Small [0\degree-30\degree] & 5.00   & 5.94 \\ 
	\hline
Medium [30\degree-60\degree] & 5.06   & 6.48 \\ 
	\hline
Large [$>$60\degree] & 6.74  & 7.96 \\ 
	\hline
\end{tabular}
\vspace{0.5em}
\caption{Performance evaluation on AFLW\cite{Zhu_2016_CVPR}. We report NMS across all visible landmarks} \label{tab:runtime}
\label{tab:aflw}
\end{center}
\end{table}

We also compare our method on the AFLW-PIFA ~\cite{Jourabloo2015} dataset with another large-pose face alignment method from Jourabloo et al.~\cite{jourabloo2016large}. 
In Table~\ref{tab:aflwpifa}, we see that our method again achieves comparable performance.
\begin{table}
\begin{center} 
\begin{tabular}{ |c|c|c| } 
 \hline
  Jourabloo et al. ~\cite{jourabloo2016large}  &Our Method\\ 
	\hline
 4.72  & 5.42 \\ 
 \hline
\end{tabular}
\vspace{0.5em}
\caption{Performance evaluation on AFLW-PIFA \cite{jourabloo2016large}. We report NMS error across the original 21 AFLW landmarks.} \label{tab:runtime}
\label{tab:aflwpifa}
\end{center}
\end{table}


Additional quantitative evaluations of our 3D model fitting and dense correspondences can be found in the supplemental materials.

\compactsubsection{Runtime}

In addition to comparably robust and accurate face alignment, our method is one to several orders of magnitude more efficient compared with other state-of-the-art methods on large pose face alignment.
Our method takes only a single iteration at test time, providing us with a large advantage in terms of efficiency. 
In Table~\ref{tab:runtime}, we summarize the runtime speed of the competing methods for large pose face alignment. Without an iterative process, it takes only 9ms on an NVIDIA Titan X GPU for our network to estimate the
dense correspondences, which is one to several orders of faster than others.
A 3D face-fitting post-process would take up to an additional 10 ms on the CPU, meaning that in total our pipeline can obtain a 3D face-fitting from an input image within 19 ms.


\begin{table}
\vspace{-5mm}
\begin{center} 
\begin{tabular}{ |c|c|c| } 
 \hline
   Jourabloo et al. ~\cite{jourabloo2016large} & Zhu et al.~\cite{Zhu_2016_CVPR} &Our Method\\ 
	\hline
  1666& 75.7 & 9.35\\ 
 \hline
\end{tabular}
\vspace{0.5em}
\caption{Comparisons of Runtimes in Milliseconds} \label{tab:runtime}
\end{center}
\vspace{-4mm}
\end{table}

%% file: conclusion.tex

\compactsubsection{Limitations}  \label{Limitations}

\begin{figure}[h]
\begin{center}
\includegraphics[width=0.24\linewidth]{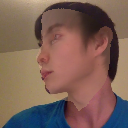}
\includegraphics[width=0.24\linewidth]{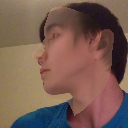}
\includegraphics[width=0.24\linewidth]{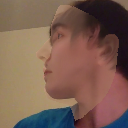}
\includegraphics[width=0.24\linewidth]{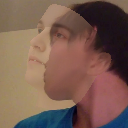}
\caption{We observe a gradual decrease in performance for pose estimation as the pose becomes more extreme and key features (i.e. nose, eye, mouth) disappear. Our algorithm completely misses on the fourth frame when the nose is completely occluded by the rest of the face. The estimated yaw angle of the previous frame is $70 \degree$.} \label{pose}
\vspace{1em}
\includegraphics[width=0.24\linewidth]{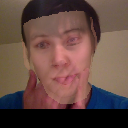}
\includegraphics[width=0.24\linewidth]{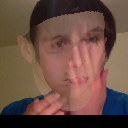}
\includegraphics[width=0.24\linewidth]{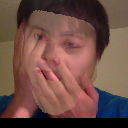}
\includegraphics[width=0.24\linewidth]{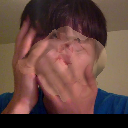}
\caption{Like with pose, we can handle the disappearance of one key feature, and performance gradually deteriorates until our algorithm completely misses when an eye, nose, and mouth are all occluded in the fourth frame.} \label{occlusion} \label{invis}
\vspace{1em}
\includegraphics[width=0.24\linewidth]{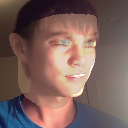}
\includegraphics[width=0.24\linewidth]{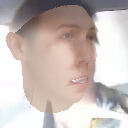}
\includegraphics[width=0.24\linewidth]{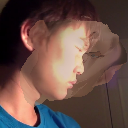}
\includegraphics[width=0.24\linewidth]{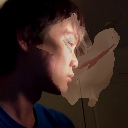}
\caption{We see that for a frontal face, we handle intense lighting conditions both from artificial and natural light quite well. However, when extreme lighting is combined with large head pose, we see our performance suffers significantly.} \label{occlusion}
\end{center}
\vspace{-2mm}
\end{figure}

We see in Figures ~\ref{pose} to ~\ref{occlusion} the limits of what our network can achieve. 
Namely, with respect to side faces and occlusion, we are robust enough to obtain good results when one key feature (i.e. face, mouth, nose) disappears, but performance drop as more features disappear due to extreme pose or occlusion.

\compactsection{Discussion}

We have presented an alternative deep learning solution for 3D facial fitting to some top performing regression based techniques~\cite{Zhu_2016_CVPR,jourabloo2016large}.
Our experiments show that it is possible to reliably estimate dense 2D facial correspondences from RGB images by training a convolutional neural network with encoder-decoder architecture
using a combination of real photographs and synthetic renderings with 3DMM variations, perturbations, and simulated occlusions and lighting. 

With the same amount of real-world training data we are 28.5\% more accurate on inner visible landmarks than Zhu et al.~\cite{Zhu_2016_CVPR} for the AFLW2000 dataset and 
14.8\% less accurate to Jourabloo et al.~\cite{jourabloo2016large} for the PIFA dataset. Generally our results can be considered comparable as other methods outperform us in certain cases (e.g their fitting among side views and wide-open mouths are slightly more accurate) while we outperform related works in other cases (e.g. we are more robust to external occlusion, illumination, and image-axis rotation).
However, our approach is significantly faster (refer to Table ~\ref{tab:runtime})
and shows increased robustness on our real test cases, such as for facial tracking in the wild, where the facial detection bounding box is not reliable and does not always provide a tight crop (see supplemental materials). 

When assessing the sparse landmark positions after a 3DMM fitting step, our approach is less accurate than some cutting-edge landmark detectors such as TVS (commercial variant of ~\cite{saragih2007nonlinear}) or ~\cite{kazemi2014one},
but we can handle extreme conditions such as large poses, challenging lighting, and occlusions. In addition to the robustness, our method is one to several orders faster than other state-of-the-art large pose face alignment methods, and is the only one that can be real-time.
Our experiments suggest that a reasonable design choice is to use our efficient and robust dense 2D flow prediction as initialization for a refined and more accurate sparse landmark detection step. 


%

Though more efficient, our current training is limited to facial shape and appearance variations from 3DMM and photos provided by Zhu et al.~\cite{Zhu_2016_CVPR}, which indicates a similar performance
to existing regression approaches. Nevertheless, since we predict 2D flows directly, we are not limited to the model space of 3DMM, and could potentially increase the dimensionality of variations, and include
more expressions, facial hair, and potentially non-realistic faces such as drawings and cartoon characters. Hence, the full capabilities of our dense correspondence approach is not fully leveraged, but 
new training data sources need to be investigated.



\vspace{-4mm}
\paragraph{Future Work.}

While we improve the state-of-the art in terms of efficiency and robustness, the presented framework is far from perfect. 
For example, although our matchability mask is able to accurately detect self-occlusion, we do not explicitly segment out external occlusions from the face region of the image due to limitations in our training data. Our accuracy is also limited by the low resolution (128x128) of the DNN input, and we would like to improve the resolution and accuracy to eliminate any need for a refinement step. 
Additionally, although we improve state-of-the-art in terms of robustness, there are still cases such as the ones listed in the Section ~\ref{Limitations} where our method fails,
and we would like to extend out method to address these limitations.

We could improve our framework by introducing more training data with accurate face segmentation ground-truth, and better ground-truth fitting of the whole head, especially the back of the head and ears, would allow us to accurately track faces with even more extreme poses where close to all the sparse landmarks that are traditionally tracked in other methods are hidden.
We will plan to explore new directions to generate
more training data with dense facial labels using both computer graphics and machine learning techniques. Recent advancements in generative adversarial networks are promising areas
for exploration. 
We could also extend the framework to directly infer 3D positions, eliminating the need to do post-hoc 3D fitting. 
If we can accurately infer dense correspondences for shapes beyond the space spanned by 3DMM, we could also model faces with more details and capture facial hair and impact general 3D reconstruction techniques such as
structure from motion and multi-view stereo.

%
%
%
\vspace{-2mm}
\paragraph{Acknowledgements}
We would like to thank Iman Sadeghi, Melanie Hamasaki, and Justin Kriebal for acting as capture models.
This
research is supported in part by Adobe, Oculus \& Facebook, Huawei, Sony, the Google Faculty
Research Award, the Okawa Foundation Research Grant, and the U.S. Army Research Laboratory
(ARL) under contract W911NF-14-D-0005. The views and conclusions contained herein are those
of the authors and should not be interpreted as necessarily representing the official policies or
endorsements, either expressed or implied, of ARL or the U.S. Government. The U.S. Government
is authorized to reproduce and distribute reprints for Governmental purpose notwithstanding any
copyright annotation thereon.

%% file: template.bbl
\begin{thebibliography}{10}\itemsep=-1pt

\bibitem{tran2016regressing}
A.~T. {a}n Tr\~{a}n, T.~Hassner, I.~Masi, and G.~Medioni.
\newblock Regressing robust and discriminative {3D} morphable models with a
  very deep neural network.
\newblock {\em arXiv}, 2016.

\bibitem{asthana2013robust}
A.~Asthana, S.~Zafeiriou, S.~Cheng, and M.~Pantic.
\newblock Robust discriminative response map fitting with constrained local
  models.
\newblock In {\em IEEE CVPR}, pages 3444--3451, 2013.

\bibitem{asthana2014incremental}
A.~Asthana, S.~Zafeiriou, S.~Cheng, and M.~Pantic.
\newblock Incremental face alignment in the wild.
\newblock In {\em IEEE CVPR}, pages 1859--1866, 2014.

\bibitem{baker2011database}
S.~Baker, D.~Scharstein, J.~Lewis, S.~Roth, M.~J. Black, and R.~Szeliski.
\newblock A database and evaluation methodology for optical flow.
\newblock {\em IJCV}, 92(1):1--31, 2011.

\bibitem{Barnes:2010}
C.~Barnes, E.~Shechtman, D.~B. Goldman, and A.~Finkelstein.
\newblock The generalized patchmatch correspondence algorithm.
\newblock In {\em IEEE ECCV}, ECCV'10, pages 29--43, Berlin, Heidelberg, 2010.
  Springer-Verlag.

\bibitem{black1996robust}
M.~J. Black and P.~Anandan.
\newblock The robust estimation of multiple motions: Parametric and
  piecewise-smooth flow fields.
\newblock {\em Computer vision and image understanding}, 63(1):75--104, 1996.

\bibitem{Blanz:1999:MMS}
V.~Blanz and T.~Vetter.
\newblock A morphable model for the synthesis of {3D} faces.
\newblock In {\em ACM SIGGRAPH}, pages 187--194, 1999.

\bibitem{Blanz:2003:FRB}
V.~Blanz and T.~Vetter.
\newblock Face recognition based on fitting a 3d morphable model.
\newblock {\em IEEE PAMI}, 25(9):1063--1074, Sept. 2003.

\bibitem{bristow2015dense}
H.~Bristow, J.~Valmadre, and S.~Lucey.
\newblock Dense semantic correspondence where every pixel is a classifier.
\newblock In {\em IEEE ICCV}, pages 4024--4031, 2015.

\bibitem{brox2011large}
T.~Brox and J.~Malik.
\newblock Large displacement optical flow: descriptor matching in variational
  motion estimation.
\newblock {\em IEEE PAMI}, 33(3):500--513, 2011.

\bibitem{burgos2013robust}
X.~P. Burgos-Artizzu, P.~Perona, and P.~Doll{\'a}r.
\newblock Robust face landmark estimation under occlusion.
\newblock In {\em IEEE ICCV}, pages 1513--1520, 2013.

\bibitem{cao2014displaced}
C.~Cao, Q.~Hou, and K.~Zhou.
\newblock Displaced dynamic expression regression for real-time facial tracking
  and animation.
\newblock {\em ACM TOG}, 33(4):43, 2014.

\bibitem{cao2014facewarehouse}
C.~Cao, Y.~Weng, S.~Zhou, Y.~Tong, and K.~Zhou.
\newblock Facewarehouse: a 3d facial expression database for visual computing.
\newblock {\em IEEE TVCG}, 20(3):413--425, 2014.

\bibitem{cao2016real}
C.~Cao, H.~Wu, Y.~Weng, T.~Shao, and K.~Zhou.
\newblock Real-time facial animation with image-based dynamic avatars.
\newblock {\em ACM TOG}, 35(4):126, 2016.

\bibitem{cao2014face}
X.~Cao, Y.~Wei, F.~Wen, and J.~Sun.
\newblock Face alignment by explicit shape regression.
\newblock {\em IEEE IJCV}, 107(2):177--190, 2014.

\bibitem{ChuRC14}
B.~Chu, S.~Romdhani, and L.~Chen.
\newblock 3d-aided face recognition robust to expression and pose variations.
\newblock In {\em IEEE CVPR}, pages 1907--1914, 2014.

\bibitem{Cootes98activeappearance}
T.~F. Cootes, G.~J. Edwards, and C.~J. Taylor.
\newblock Active appearance models.
\newblock {\em IEEE PAMI}, 23(6):681--685, 2001.

\bibitem{cootes2002view}
T.~F. Cootes, G.~V. Wheeler, K.~N. Walker, and C.~J. Taylor.
\newblock View-based active appearance models.
\newblock {\em Image and vision computing}, 20(9):657--664, 2002.

\bibitem{Cristinacce:2008:AFL}
D.~Cristinacce and T.~Cootes.
\newblock Automatic feature localisation with constrained local models.
\newblock {\em Pattern Recogn.}, 41(10):3054--3067, 2008.

\bibitem{dosovitskiy2015flownet}
A.~Dosovitskiy, P.~Fischer, E.~Ilg, P.~Hausser, C.~Hazirbas, V.~Golkov,
  P.~van~der Smagt, D.~Cremers, and T.~Brox.
\newblock Flownet: Learning optical flow with convolutional networks.
\newblock In {\em IEEE ICCV}, pages 2758--2766, 2015.

\bibitem{ghiasi2014occlusion}
G.~Ghiasi and C.~C. Fowlkes.
\newblock Occlusion coherence: Localizing occluded faces with a hierarchical
  deformable part model.
\newblock In {\em IEEE CVPR}, pages 1899--1906, 2014.

\bibitem{guler2016densereg}
R.~A. G{\"u}ler, G.~Trigeorgis, E.~Antonakos, P.~Snape, S.~Zafeiriou, and
  I.~Kokkinos.
\newblock Densereg: Fully convolutional dense shape regression in-the-wild.
\newblock {\em arXiv preprint arXiv:1612.01202}, 2016.

\bibitem{horn1981determining}
B.~K. Horn and B.~G. Schunck.
\newblock Determining optical flow.
\newblock {\em Artificial intelligence}, 17(1-3):185--203, 1981.

\bibitem{HuYCDCKR16}
G.~Hu, F.~Yan, C.~Chan, W.~Deng, W.~J. Christmas, J.~Kittler, and N.~M.
  Robertson.
\newblock Face recognition using a unified 3d morphable model.
\newblock In {\em IEEE ECCV}, pages 73--89, 2016.

\bibitem{HuR16}
P.~Hu and D.~Ramanan.
\newblock Finding tiny faces.
\newblock {\em CoRR}, abs/1612.04402, 2016.

\bibitem{Ichim:2015:DAC}
A.~E. Ichim, S.~Bouaziz, and M.~Pauly.
\newblock Dynamic 3d avatar creation from hand-held video input.
\newblock In {\em ACM SIGGRAPH}, pages 45:1--45:14, 2015.

\bibitem{ilg2016flownet}
E.~Ilg, N.~Mayer, T.~Saikia, M.~Keuper, A.~Dosovitskiy, and T.~Brox.
\newblock Flownet 2.0: Evolution of optical flow estimation with deep networks.
\newblock {\em arXiv}, 2016.

\bibitem{jeni2015dense}
L.~A. Jeni, J.~F. Cohn, and T.~Kanade.
\newblock Dense 3d face alignment from 2d videos in real-time, 2015.

\bibitem{jia2014structured}
X.~Jia, H.~Yang, A.~Lin, K.-P. Chan, and I.~Patras.
\newblock Structured semi-supervised forest for facial landmarks localization
  with face mask reasoning.
\newblock In {\em BMVC}, 2014.

\bibitem{Jourabloo2015}
A.~Jourabloo and X.~Liu.
\newblock Pose-invariant 3d face alignment.
\newblock In {\em IEEE ICCV}, pages 3694--3702, Dec 2015.

\bibitem{jourabloo2016large}
A.~Jourabloo and X.~Liu.
\newblock Large-pose face alignment via cnn-based dense 3d model fitting.
\newblock In {\em IEEE CVPR}, pages 4188--4196, 2016.

\bibitem{kazemi2014one}
V.~Kazemi and J.~Sullivan.
\newblock One millisecond face alignment with an ensemble of regression trees.
\newblock In {\em IEEE CVPR}, pages 1867--1874, 2014.

\bibitem{kim2013deformable}
J.~Kim, C.~Liu, F.~Sha, and K.~Grauman.
\newblock Deformable spatial pyramid matching for fast dense correspondences.
\newblock In {\em IEEE CVPR}, pages 2307--2314, 2013.

\bibitem{tugraz:icg:lrs:koestinger11b}
M.~Koestinger, P.~Wohlhart, P.~M. Roth, and H.~Bischof.
\newblock Annotated facial landmarks in the wild: A large-scale, real-world
  database for facial landmark localization.
\newblock In {\em IEEE Int. Workshop on Benchmarking Facial Image Analysis
  Technologies}, 2011.

\bibitem{laine2016facial}
S.~Laine, T.~Karras, T.~Aila, A.~Herva, S.~Saito, R.~Yu, H.~Li, and
  J.~Lehtinen.
\newblock Facial performance capture with deep neural networks.
\newblock {\em arXiv preprint arXiv:1609.06536}, 2016.

\bibitem{lee2015face}
D.~Lee, H.~Park, and C.~D. Yoo.
\newblock Face alignment using cascade gaussian process regression trees.
\newblock In {\em IEEE CVPR}, pages 4204--4212, 2015.

\bibitem{li2015facial}
H.~Li, L.~Trutoiu, K.~Olszewski, L.~Wei, T.~Trutna, P.-L. Hsieh, A.~Nicholls,
  and C.~Ma.
\newblock Facial performance sensing head-mounted display.
\newblock {\em ACM Transactions on Graphics (Proceedings SIGGRAPH 2015)},
  34(4), July 2015.

\bibitem{Lin2014}
T.-Y. Lin, M.~Maire, S.~Belongie, J.~Hays, P.~Perona, D.~Ramanan,
  P.~Doll{\'a}r, and C.~L. Zitnick.
\newblock Microsoft coco: Common objects in context.
\newblock In D.~Fleet, T.~Pajdla, B.~Schiele, and T.~Tuytelaars, editors, {\em
  IEEE ECCV}, pages 740--755, Cham, 2014.

\bibitem{liu2011sift}
C.~Liu, J.~Yuen, and A.~Torralba.
\newblock Sift flow: Dense correspondence across scenes and its applications.
\newblock {\em IEEE PAMI}, 33(5):978--994, 2011.

\bibitem{liu2016joint}
F.~Liu, D.~Zeng, Q.~Zhao, and X.~Liu.
\newblock Joint face alignment and 3d face reconstruction.
\newblock In {\em IEEE ECCV}, pages 545--560, 2016.

\bibitem{Ma2015}
D.~S. Ma, J.~Correll, and B.~Wittenbrink.
\newblock The chicago face database: A free stimulus set of faces and norming
  data.
\newblock {\em Behavior Research Methods}, 47(4):1122--1135, 2015.

\bibitem{matthews2004active}
I.~Matthews and S.~Baker.
\newblock Active appearance models revisited.
\newblock {\em IEEE IJCV}, 60(2):135--164, 2004.

\bibitem{olszewski2016high}
K.~Olszewski, J.~J. Lim, S.~Saito, and H.~Li.
\newblock High-fidelity facial and speech animation for vr hmds.
\newblock {\em ACM Transactions on Graphics (Proceedings SIGGRAPH Asia 2016)},
  35(6), December 2016.

\bibitem{PaysanKARV09}
P.~Paysan, R.~Knothe, B.~Amberg, S.~Romdhani, and T.~Vetter.
\newblock A 3d face model for pose and illumination invariant face recognition.
\newblock In S.~Tubaro and J.-L. Dugelay, editors, {\em AVSS}, pages 296--301.
  IEEE Computer Society, 2009.

\bibitem{peng2016recurrent}
X.~Peng, R.~S. Feris, X.~Wang, and D.~N. Metaxas.
\newblock A recurrent encoder-decoder network for sequential face alignment.
\newblock In {\em IEEE ECCV}, pages 38--56, 2016.

\bibitem{ren2014face}
S.~Ren, X.~Cao, Y.~Wei, and J.~Sun.
\newblock Face alignment at 3000 fps via regressing local binary features.
\newblock In {\em IEEE CVPR}, pages 1685--1692, 2014.

\bibitem{romdhani2005estimating}
S.~Romdhani and T.~Vetter.
\newblock Estimating 3d shape and texture using pixel intensity, edges,
  specular highlights, texture constraints and a prior.
\newblock In {\em IEEE CVPR}, volume~2, pages 986--993, 2005.

\bibitem{sagonas2013300}
C.~Sagonas, G.~Tzimiropoulos, S.~Zafeiriou, and M.~Pantic.
\newblock 300 faces in-the-wild challenge: The first facial landmark
  localization challenge.
\newblock In {\em Proceedings of the IEEE International Conference on Computer
  Vision Workshops}, pages 397--403, 2013.

\bibitem{saito2016realtime}
S.~Saito, T.~Li, and H.~Li.
\newblock Real-time facial segmentation and performance capture from rgb input.
\newblock In {\em IEEE ECCV}, 2016.

\bibitem{SaitoWHNL16}
S.~Saito, L.~Wei, L.~Hu, K.~Nagano, and H.~Li.
\newblock Photorealistic facial texture inference using deep neural networks.
\newblock In {\em IEEE CVPR}, 2016.

\bibitem{saragih2007nonlinear}
J.~Saragih and R.~Goecke.
\newblock A nonlinear discriminative approach to aam fitting.
\newblock In {\em IEEE ICCV}, pages 1--8, 2007.

\bibitem{Saragih:2011:DMF}
J.~M. Saragih, S.~Lucey, and J.~F. Cohn.
\newblock Deformable model fitting by regularized landmark mean-shift.
\newblock {\em IJCV}, 91(2):200--215, 2011.

\bibitem{sun2013deep}
Y.~Sun, X.~Wang, and X.~Tang.
\newblock Deep convolutional network cascade for facial point detection.
\newblock In {\em IEEE CVPR}, pages 3476--3483, 2013.

\bibitem{taniai2016joint}
T.~Taniai, S.~N. Sinha, and Y.~Sato.
\newblock Joint recovery of dense correspondence and cosegmentation in two
  images.
\newblock In {\em IEEE CVPR}, pages 4246--4255, 2016.

\bibitem{thies2016face}
J.~Thies, M.~Zollh{\"o}fer, M.~Stamminger, C.~Theobalt, and M.~Nie{\ss}ner.
\newblock Face2face: Real-time face capture and reenactment of rgb videos.
\newblock In {\em IEEE CVPR}, 2016.

\bibitem{tzimiropoulos2013optimization}
G.~Tzimiropoulos and M.~Pantic.
\newblock Optimization problems for fast aam fitting in-the-wild.
\newblock In {\em IEEE ICCV}, pages 593--600, 2013.

\bibitem{valstar2010facial}
M.~Valstar, B.~Martinez, X.~Binefa, and M.~Pantic.
\newblock Facial point detection using boosted regression and graph models.
\newblock In {\em IEEE CVPR}, pages 2729--2736, 2010.

\bibitem{weinzaepfel2013deepflow}
P.~Weinzaepfel, J.~Revaud, Z.~Harchaoui, and C.~Schmid.
\newblock Deepflow: Large displacement optical flow with deep matching.
\newblock In {\em IEEE ICCV}, pages 1385--1392, 2013.

\bibitem{xiong2013supervised}
X.~Xiong and F.~De~la Torre.
\newblock Supervised descent method and its applications to face alignment.
\newblock In {\em IEEE CVPR}, pages 532--539, 2013.

\bibitem{Xiong2015}
X.~Xiong and F.~D. la~Torre.
\newblock Global supervised descent method.
\newblock In {\em IEEE CVPR}, pages 2664--2673, June 2015.

\bibitem{yang2015robust}
H.~Yang, X.~He, X.~Jia, and I.~Patras.
\newblock Robust face alignment under occlusion via regional predictive power
  estimation.
\newblock {\em IEEE Trans. on Image Proc.}, 2015.

\bibitem{yu2013pose}
X.~Yu, J.~Huang, S.~Zhang, W.~Yan, and D.~N. Metaxas.
\newblock Pose-free facial landmark fitting via optimized part mixtures and
  cascaded deformable shape model.
\newblock In {\em IEEE ICCV}, pages 1944--1951, 2013.

\bibitem{zhang2014facial}
Z.~Zhang, P.~Luo, C.~C. Loy, and X.~Tang.
\newblock Facial landmark detection by deep multi-task learning.
\newblock In {\em IEEE ECCV}, pages 94--108, 2014.

\bibitem{zhou2013extensive}
E.~Zhou, H.~Fan, Z.~Cao, Y.~Jiang, and Q.~Yin.
\newblock Extensive facial landmark localization with coarse-to-fine
  convolutional network cascade.
\newblock In {\em IEEE ICCV Workshops}, pages 386--391, 2013.

\bibitem{zhou2015flowweb}
T.~Zhou, Y.~Jae~Lee, S.~X. Yu, and A.~A. Efros.
\newblock Flowweb: Joint image set alignment by weaving consistent, pixel-wise
  correspondences.
\newblock In {\em IEEE CVPR}, pages 1191--1200, 2015.

\bibitem{Zhou_2016_CVPR}
T.~Zhou, P.~Krahenbuhl, M.~Aubry, Q.~Huang, and A.~A. Efros.
\newblock Learning dense correspondence via 3d-guided cycle consistency.
\newblock In {\em IEEE CVPR}, June 2016.

\bibitem{zhu2015face}
S.~Zhu, C.~Li, C.~C. Loy, and X.~Tang.
\newblock Face alignment by coarse-to-fine shape searching.
\newblock In {\em IEEE CVPR}, pages 4998--5006, 2015.

\bibitem{Zhu_2016_CVPR}
X.~Zhu, Z.~Lei, X.~Liu, H.~Shi, and S.~Z. Li.
\newblock Face alignment across large poses: A 3d solution.
\newblock In {\em IEEE CVPR}, June 2016.

\bibitem{zhu2015high}
X.~Zhu, Z.~Lei, J.~Yan, D.~Yi, and S.~Z. Li.
\newblock High-fidelity pose and expression normalization for face recognition
  in the wild.
\newblock In {\em IEEE CVPR}, pages 787--796, 2015.

\bibitem{zhu2012face}
X.~Zhu and D.~Ramanan.
\newblock Face detection, pose estimation, and landmark localization in the
  wild.
\newblock In {\em IEEE CVPR}, pages 2879--2886, 2012.

\end{thebibliography}
